\documentclass{Interspeech}

\interspeechcameraready

\title{Feature-based analysis of oral narratives from Afrikaans and isiXhosa children}

\author[affiliation={1}]{Emma}{Sharratt}
\author[affiliation={2}]{Annelien}{Smith}
\author[affiliation={1}]{Retief}{Louw}
\author[affiliation={2}]{Daleen}{Klop}
\author[affiliation={1}]{Febe}{de Wet}
\author[affiliation={1}]{Herman}{Kamper}

\affiliation{Electrical and Electronic Engineering}{Stellenbosch University}{South Africa}
\affiliation{Speech, Language and Hearing Therapy}{Stellenbosch University}{South Africa}

\email{emsharratt@gmail.com} 

\usepackage[prependcaption,textsize=scriptsize]{todonotes}

\usepackage{comment}
\definecolor{Ecolor}{HTML}{74a800}

\usepackage{cite}
\usepackage{balance}
\usepackage{booktabs}
\usepackage{multirow}
\usepackage{tabularx}
\newcommand{\mytable}{
	\centering
	\renewcommand{\arraystretch}{1.2}
}
\newcolumntype{C}{>{\centering\arraybackslash}X}
\newcolumntype{L}{>{\raggedright\arraybackslash}X}
\newcolumntype{R}{>{\raggedleft\arraybackslash}X}
\newcolumntype{P}[1]{>{\raggedright\arraybackslash}p{#1}}
\usepackage{siunitx}
\usepackage{etoolbox}                           
\newcommand{\ubold}{\fontseries{b}\selectfont}  
\robustify\ubold                                

\graphicspath{{fig/}}
\usepackage{pgfplots}
\pgfplotsset{compat=1.18}
\usepackage{graphicx}
\usepackage{subcaption} 
\usepackage[utf8]{inputenc}
\usepackage{tcolorbox}

\usepackage{microtype}

\usepackage[prependcaption,textsize=scriptsize]{todonotes}
\definecolor{mycolor}{HTML}{FF6600}
\definecolor{coolcolor}{HTML}{26209e}
\definecolor{plancolor}{HTML}{aaa6ed}
\definecolor{Ecolor}{HTML}{74a800}

\begin{document}
\maketitle

\begin{abstract}
Oral narrative skills are strong predictors of later literacy development. This study examines the features of oral narratives from children who were identified by experts as requiring intervention. Using simple machine learning methods, we analyse recorded stories from four- and five-year-old Afrikaans- and isiXhosa-speaking children. Consistent with prior research, we identify lexical diversity (unique words) and length-based features (mean utterance length) as indicators of typical development, but features like articulation rate prove less informative. 
Despite cross-linguistic variation in part-of-speech patterns, the use of specific verbs and auxiliaries associated with goal-directed storytelling is correlated with a reduced likelihood of requiring intervention.
Our analysis of two linguistically distinct languages reveals both language-specific and shared predictors of narrative proficiency, with implications for early assessment in multilingual contexts.
\end{abstract}

\section{Introduction}
\label{sec:intro}
Even before children learn to read, the ability to tell and understand stories is a key developmental skill.  
Oral storytelling predicts later reading proficiency~\cite{hayward_schneider_etal2009, hjetland_brinchmann_etal2020, babayigit_roulstone_etal2021} and narrative assessment can therefore help identify delayed development 
early on~\cite{dickinson_mccabe_etal2003, schick_melzi2010, reese_leyva_etal2010, oakhill_cain2012, gardner-neblett_iruka2015}.  
Yet, in many parts of the world large classroom sizes make narrative assessment impossible.
Observational assessments, where teachers rely on intuition, are often inaccurate~\cite{shermis_burstein2003,mozer_miratrix_etal2021,mozer_miratrix_etal2023}.
Our goal is to identify features that characterise age-appropriate oral narratives among preschool children in low-resource language settings.  
These findings aim to inform the development of more effective oral narrative assessment methods to support early identification of language difficulties in pre-literate children.

We focus on Afrikaans- and isiXhosa-speaking children aged four to five from low-income communities in South Africa, where large class sizes and low literacy rates are widespread~\cite{roux_van_staden_etal2023}.
We use a dataset where children told stories
based on a picture sequence (Figure~\ref{fig:MAIN_story}) and answered comprehension questions.
Education experts then assessed whether intervention was needed.
We do a text-based analysis of the narrative transcripts, using logistic regression to predict which children may need support.
In contrast to our parallel work~\cite{louw_sharratt_etal2025}, our goal here is not an accurate system for automatically identifying at-risk children, but rather to characterise the types of features that are indicative of risk using an interpretable machine learning approach.
Using permutation feature importance (PFI)~\cite{altmann_tolosi_etal2010} and exploratory data analysis, we perform quantitative analyses
to compare the informativeness of different features.
We then compare these quantitative findings to a qualitative analysis based on traditional measures of storytelling development from speech therapy.

We consider a range of features.
As in other studies~\cite{uccelli_paez2007, chan_chen_etal2023}, we find that indicative 
features include those measuring lexical diversity (unique words), productivity
(mean utterance length) and syntactic complexity (measured using a readability index).
In contrast to other studies~\cite{amster1984, mahr_soriano_etal2021}, we find that speech production features such as articulation rate are of little importance.
We also consider grammatical-level indicators (part-of-speech counts) and specific keywords. 
Although keywords are 
unique to each language, we find that specific verbs and auxiliaries linked to goal-directed sentences are associated with a lower likelihood of 
intervention in both Afrikaans and isiXhosa.

In contrast to other studies that look at single languages in isolation~\cite{miratrix_ackerman2016, mozer_miratrix_etal2021, mozer_miratrix_etal2023},
our contribution is an analysis 
of similar oral narratives from two distinct languages.
This allows us to identify features that are important to individual communities and also to identify general cross-lingual characteristics of children's oral narratives.
We hope that this quantitative study will inform further work on oral narrative assessments.

\section{Data}
\label{sec:data}

\begin{figure}[!t]
    \centering
    \includegraphics[width=0.67\linewidth]{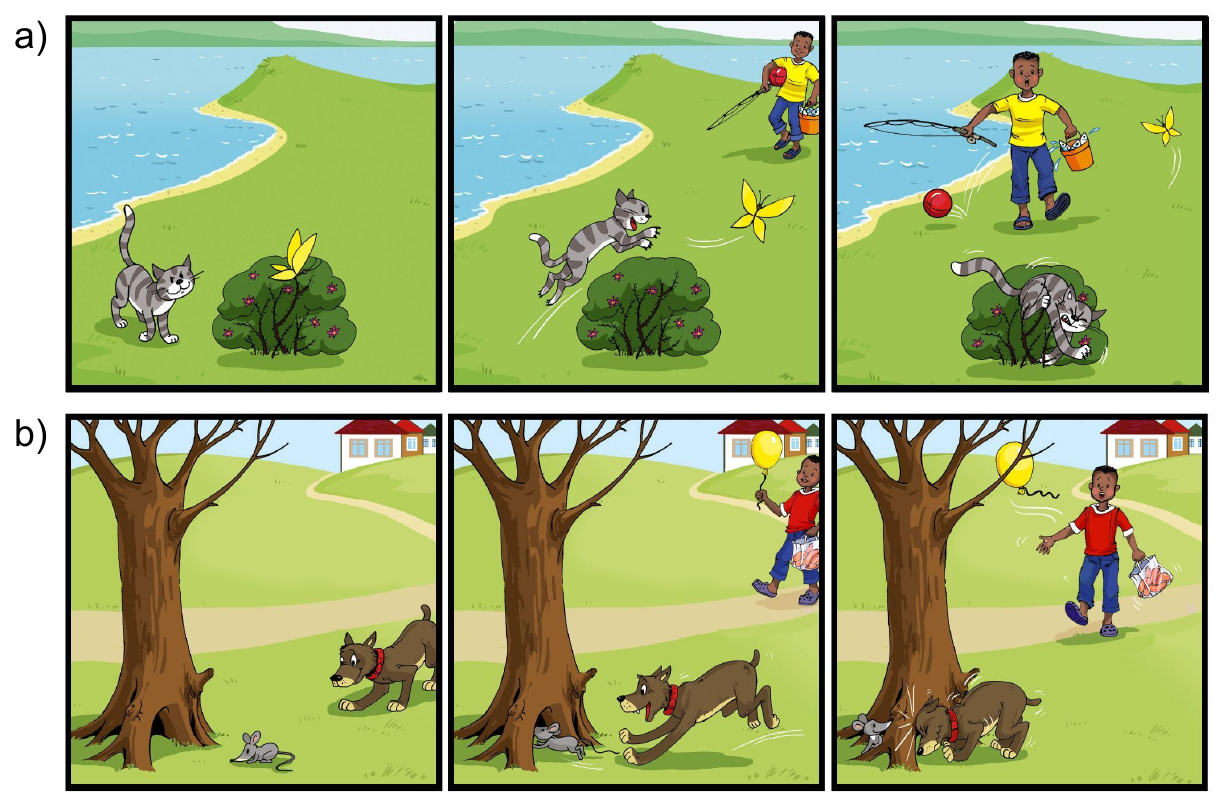}
    \caption{Three images from the six-picture cat (a) and dog (b) stories used in the MAIN protocol to elicit child narratives.}
    \label{fig:MAIN_story}
\end{figure} 

We use data from a speech and language therapy study that involved children aged four to five from lower socioeconomic backgrounds in South Africa \cite{smith2023, cain_ocarroll_etal2024}.
Each child was shown a six-picture sequence and asked to tell a story in their home language of either Afrikaans or isiXhosa.\footnote{isiXhosa and Afrikaans, with 8 million and 7.2 million speakers respectively, are two of South Africa's 12 official languages.
isiXhosa is a Southern Bantu language, while Afrikaans is a West Germanic language derived from Dutch. 
Both languages use the Latin alphabet.}  
One of two story sequences was used: a cat story or a dog story (Figure~\ref{fig:MAIN_story}).
Comprehension was subsequently assessed through questions about the story content. 
The children were evaluated using 
the multilingual assessment instrument for narratives~(MAIN)~\cite{gagarina_klop_etal2019}.
This tool was 
standardised for South African populations, accounting for cultural biases~\cite{klop_visser2020}.
Assessments were conducted by trained assessors experienced in early childhood development. 
Based on the combination of stories and comprehension answers, experts finally assigned a binary label for whether intervention is required or not~\cite{westby2005}.
We call this label \textit{requires intervention}~(RI), with a value of 1 if intervention is needed and 0 if not.
Children needing additional language support typically struggle to formulate narratives with clear goals, attempts and outcomes, e.g., \textit{She wants it. She jumps. She misses it}.

We divide the data into training (approximately 200 children), development (38 children) and test (28 children) sets for each language, with no speaker overlap between sets.
Cat and dog stories appear in all sets. 
The children's speech was recorded using Samsung Galaxy Tab A7 Lite 8.7 tablets and Logitech H111 headsets.
Fieldworkers provided an audio and transcription file for each child. 
Transcriptions were manually aligned to audio using the Praat toolkit~\cite{boersma2007}.
This resulted in a total 
of about 5~hours of active child speech for each language.
The data shows clear cross-linguistic differences.
For example, isiXhosa has an agglutinative orthography, typical of Bantu languages.
Its morphological richness leads to a larger vocabulary~\cite{mzamo_helberg_etal2015} with over 3,900 unique word types in the isiXhosa data compared to less than 1,150 
in the Afrikaans transcriptions.

\section{Methodology}
\label{sec:method}

We aim to identify which narrative features are most associated with a child requiring intervention. 
To do this, we train a linear model to predict intervention 
and use statistical methods to assess feature importance.

\subsection{Logistic regression}

As our base machine learning model, we
use logistic regression to model the combined influence of multiple features, helping to control for potential confounding effects. 
While linear models are well-suited for working with limited data, they remain susceptible to overfitting, especially in 
cases such as ours where there are only 200 training examples.
We address this in two ways. First, we use
\texttt{scikit-learn}’s \texttt{LIBLINEAR} solver with L2 regularisation~\cite{pedregosa_varoquaux_etal2011},
recommended for smaller datasets.

Second, instead of training a single model on all features at once, we 
group specific feature types. Each group is designed to show the relative importance of particular features to one another. 
For example, we want to see the relative importance of verb count compared to nouns, so these are placed in one group.
We consider three groups: verbal language proficiency features, grammatical features and keywords per language.
Three distinct logistic regression models are therefore trained.

Because the different features use different scales, we cannot use the weights of the resulting logistic regression model to measure feature importance directly.
One remedy would be to normalise the features beforehand, but this resulted in substantially worse development performance. 
We therefore retain the raw feature values and turn to the following approach. 

\subsection{Permutation feature importance (PFI)}
\label{sec:permutation_importance}

Permutation feature importance (PFI)
quantifies how much each feature contributes to a model’s performance by measuring the drop in accuracy when that feature is randomly shuffled.
A greater drop indicates greater feature importance~\cite{kaneko2022}.
We use \texttt{scikit-learn}’s implementation~\cite{pedregosa_varoquaux_etal2011}, with balanced accuracy as the evaluation metric~(Sec.~\ref{sec:metric}).
First, the 
model’s baseline performance 
$s$ is computed on the original dataset $\mathcal{D}$.
Each feature $j$ is then considered in turn. 
Column $j$ in $\mathcal{D}$ is shuffled to produce a corrupted 
dataset $\mathcal{\tilde{D}}_{r,j}$.
The performance $s_{r,j}$ on this 
dataset is calculated for repetition~$r$ and feature~$j$.
This is repeated $R$ times. 
Finally, the importance of feature $j$ is obtained
as the average drop in score compared to using the full model:~$s~-~\frac{1}{R}~\sum_{r=1}^{R} s_{r,j}$.
We use $R=100$ repetitions.

We visualise feature importance using permutation plots. 
E.g., in Figure~\ref{fig:PFI_speech_features}, the $y$-axis lists features and the $x$-axis measures importance, giving the drop in balanced accuracy across $R$ repetitions when 
corrupting the feature. 
While PFI quantifies importance, it does not show whether higher or lower values increase the likelihood of predicting intervention. 
I.e., a feature may have a large impact on accuracy (high $x$ value) without revealing the direction of influence.
To address this, we use colour:~red for positive model coefficients (predicts intervention), blue for negative (predicts no intervention). 
For example, Figure~\ref{fig:PFI_speech_features}a shows \textit{unique words} in blue, meaning more unique words are linked to typical development. In contrast, \textit{articulation rate} is red, suggesting faster speech predicts intervention. However, its low $x$ value also indicates limited importance.

\subsection{Evaluation metrics}
\label{sec:metric}

Children requiring intervention are labelled as positive, $\text{RI} = 1$, and those who do not need intervention are labelled as negative, $\text{RI} = 0$.
A true positive is therefore a correct prediction that intervention is required.
We use balanced accuracy as our primary evaluation metric: $\frac{1}{2} \left( \frac{\text{TP}}{\text{TP} + \text{FN}} + \frac{\text{TN}}{\text{TN} + \text{FP}} \right)$.
Balanced accuracy accounts for class imbalance. Furthermore, it explicitly incorporates true negative predictions~\cite{van_zyl_engelbrecht2025} which are not accounted for in metrics like precision, recall and F1.
We 
also report F1 for a full picture.

\section{Feature analysis and results}
\label{sec:results}
We use three models for our feature analysis, each using a distinct feature group: verbal language proficiency features, grammatical features and 
frequently occurring keywords.
Before we turn to our main question of the relative importance of different features within each group,
we report base model performance.

\subsection{Base model performance}
\label{sec:model_perform}
\begin{table}[!b]
    \caption{
    F1 and balanced accuracy (\%) for base models on train and development sets, with true 
    RI proportions per set.
    }
    \eightpt
    \label{tbl:model_perform}
    \mytable
    \begin{tabularx}{\linewidth}{@{}lCCCCCCCC@{}}
        \toprule
        & \multicolumn{4}{c}{Afrikaans} & \multicolumn{4}{c}{isiXhosa}  \\
         & \multicolumn{2}{c}{Train} & \multicolumn{2}{c}{Dev} & \multicolumn{2}{c}{Train} & \multicolumn{2}{c}{Dev} \\
         & \multicolumn{2}{c}{(36\% RI)} & \multicolumn{2}{c}{(32\% RI)} & \multicolumn{2}{c}{(47\% RI)} & \multicolumn{2}{c}{(53\% RI)} \\
         \cmidrule(l){2-3} \cmidrule(l){4-5} \cmidrule(l){6-7} \cmidrule(l){8-9}
        Model & F1 & Acc. & F1 & Acc. & F1 & Acc. & F1 & Acc. \\
        \midrule
        Proficiency & 64 &  71 & 64 & 74 & 69 & 69 & 62 & 61 \\
        Grammatical 
        & 62 & 69 & 52 & 64 & 71 & 71 & 61 & 64 \\
        Keywords & 70 & 76 & 55 & 67 & 79 & 79 & 73 & 71 \\
        All features & 77 & 82 & 55 & 67 & 84 & 84 & 61 & 63 \\
        \bottomrule
    \end{tabularx}
    \end{table}
Table~\ref{tbl:model_perform} summarises the base systems' accuracy on training and development sets for both languages. 
We also show the class distribution, e.g., in the Afrikaans training set, positive RI labels make up 36\% of the targets.
We perform our analyses on the training data.
We can do this because our goal is not to evaluate absolute performance but rather to understand what the data reveals about feature relevance.
Moreover, because it is a larger set, it 
yields more robust results.
We use the training set's balanced accuracies in Table~\ref{tbl:model_perform} 
as the reference scores (Sec.~\ref{sec:permutation_importance}) for assessing feature importance in the analyses that follow.

\subsection{Which verbal proficiency features are important?}
\label{sec:verbal_prof}

\begin{figure}[!t]
  \centering
  \begin{subfigure}[b]{0.475\linewidth}
    \includegraphics[width=\linewidth]{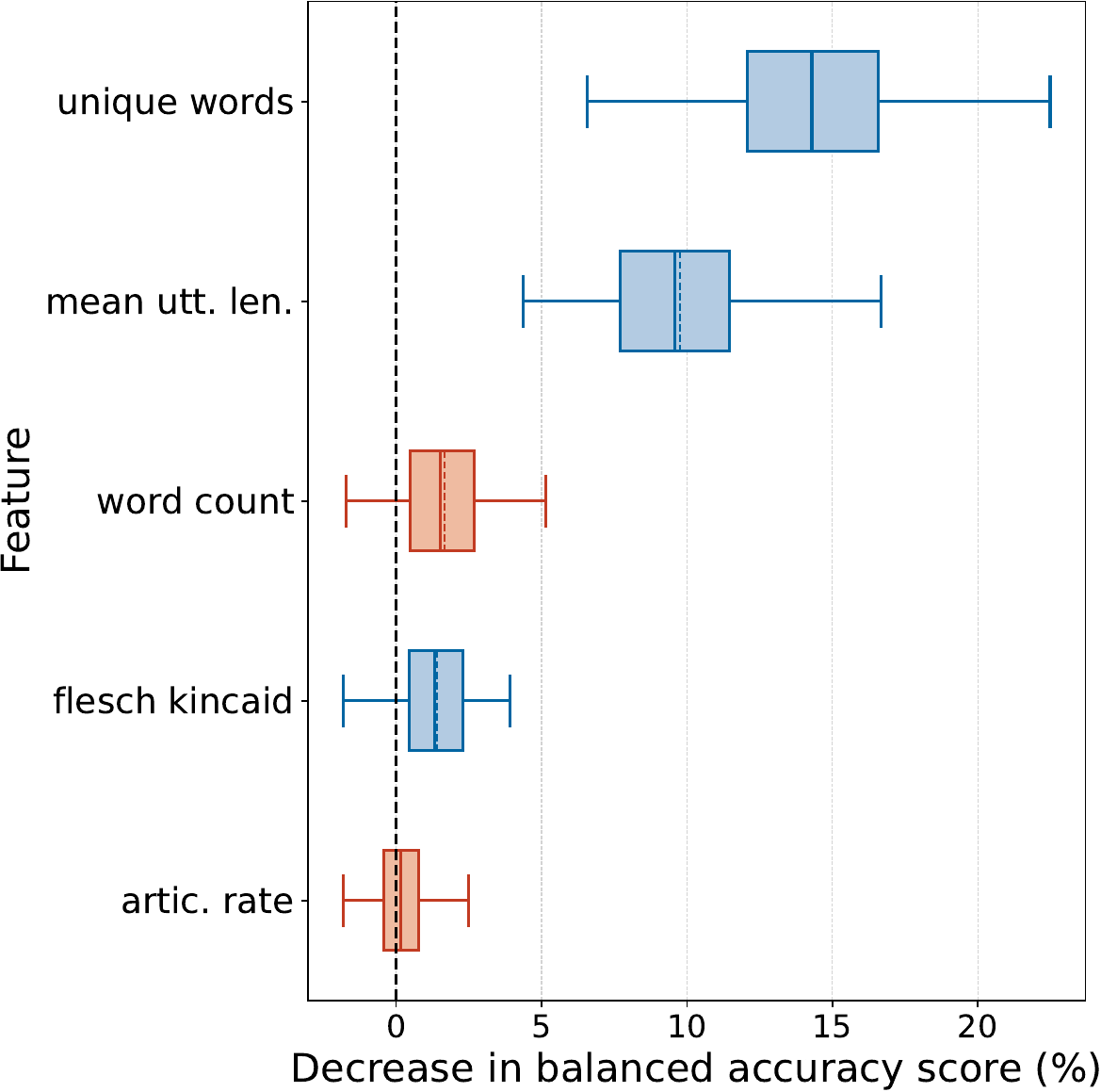}
    \caption{Afrikaans}
    \label{fig:PFI_af_speech_features}
  \end{subfigure}
  \begin{subfigure}[b]{0.46\linewidth}
    \includegraphics[width=\linewidth]{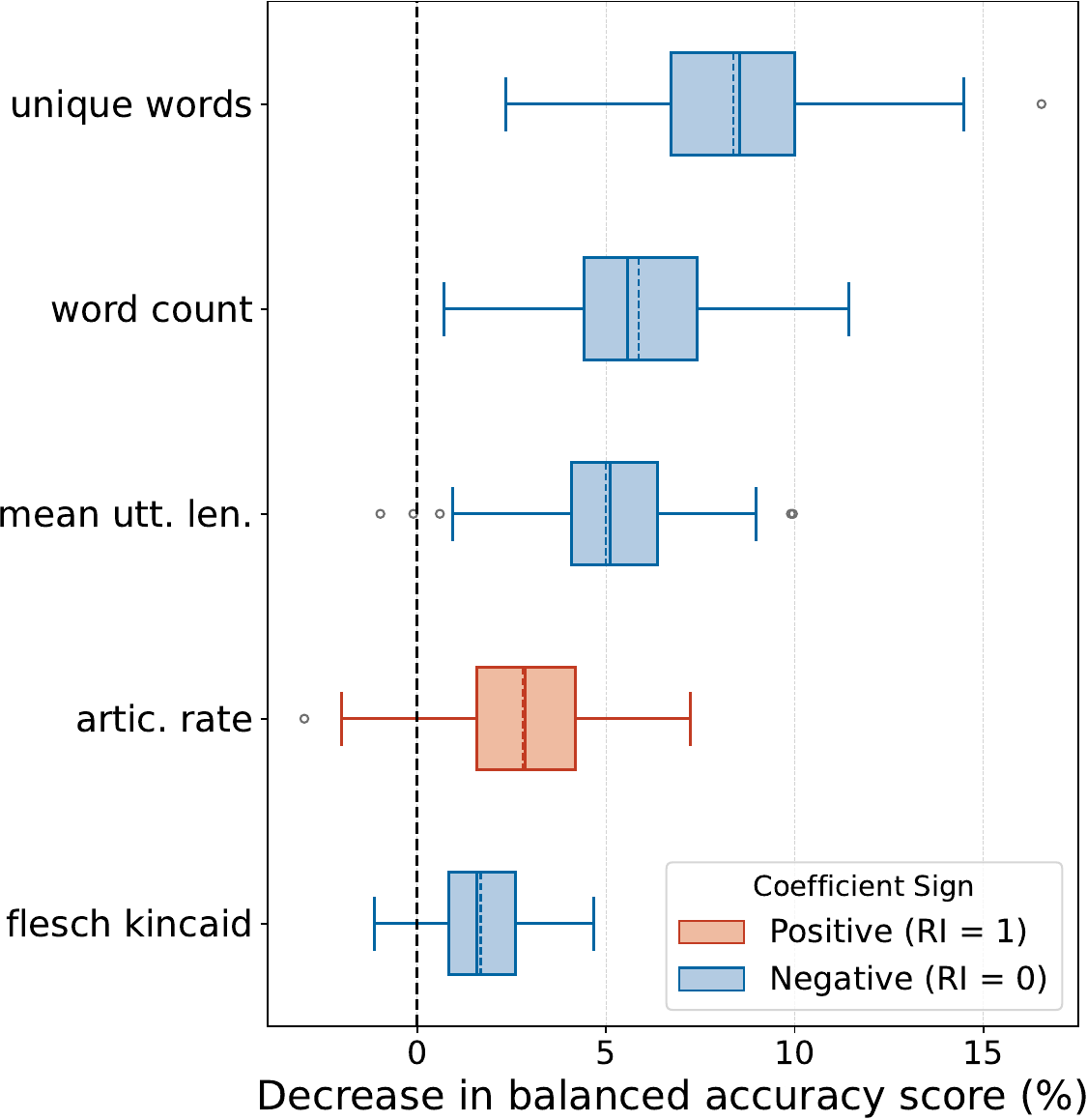}
    \caption{isiXhosa}
    \label{fig:PFI_xh_speech_features}
  \end{subfigure}
  \caption{PFI 
  for verbal language proficiency features, showing the drop in balanced accuracy when a given feature is corrupted.}
  \label{fig:PFI_speech_features}
\end{figure}

\hspace{\parindent}\textbf{Setup:}
We start by considering the relative importance of verbal language proficiency features.
These include the total number of word tokens and types extracted from the normalised transcriptions; mean utterance length (seconds) and articulation rate (characters per second)\footnote{Both languages have regular grapheme-to-phoneme mappings, making characters per second a good proxy for phonemes per second.} derived from the recordings; and the Flesch–Kincaid readability score\footnote{In development experiments, we also looked at other readability indices, but Flesch–Kincaid gave the most reliable results.} which estimates linguistic complexity based on sentence length and syllable count~\cite{flesch1948, kincaid1975}.

\textbf{Quantitative analysis:}
As seen in the top bar of Figure~\ref{fig:PFI_speech_features}, unique word count has the highest feature importance for both languages, suggesting a larger vocabulary 
corresponds with predicting satisfactory development (non-RI). 
Looking at Figure~\ref{fig:PFI_speech_features}, we see that mean utterance length is the third most important proficiency feature for isiXhosa and the second for Afrikaans.
In both languages, longer utterances are 
linked to typical development.
Turning to the word count feature, Figure~\ref{fig:PFI_xh_speech_features} shows that the number of words produced is a key feature in 
isiXhosa, 
with higher counts associated with non-RI classification (blue).
Surprisingly, for the Afrikaans model, higher word count is associated with intervention (red).
But, its low feature importance 
suggests that it plays a limited role in classifying whether intervention is required in  Afrikaans.
Finally, articulation rate is of low importance for both languages, with an unexpected trend that faster speech is associated with requiring intervention.

\begin{figure}[!t]
  \centering
  \begin{subfigure}[b]{0.29\linewidth}
    \includegraphics[width=\linewidth]{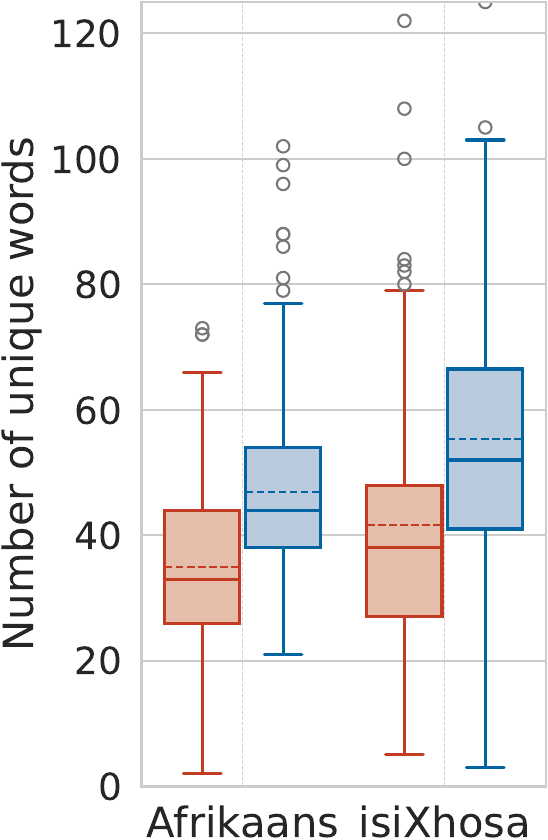}
     \caption{Unique words}
    \label{fig:unique_words_box}
  \end{subfigure}
  \begin{subfigure}[b]{0.29\linewidth}
    \includegraphics[width=\linewidth]{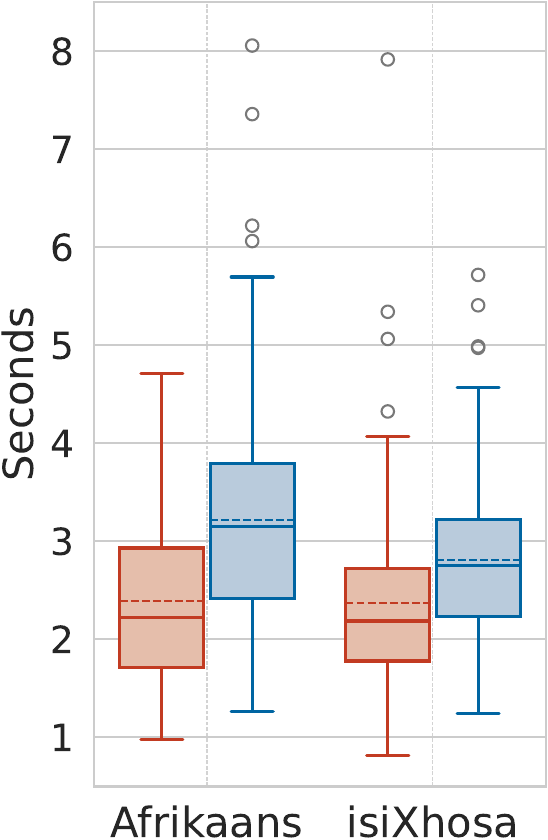}
    \caption{Mean utt. length}
    \label{fig:mean_utt_box}
  \end{subfigure}
  \begin{subfigure}[b]{0.29\linewidth}
    \includegraphics[width=\linewidth]{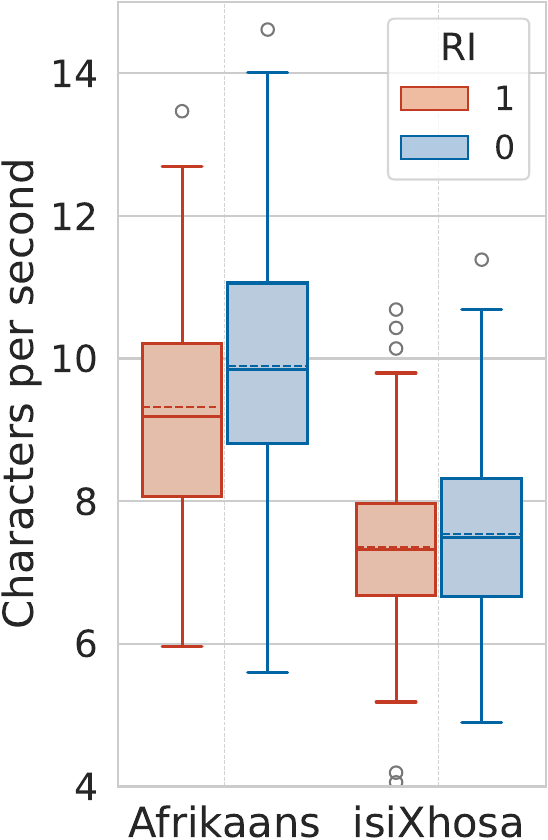}
    \caption{Articulation rate}
    \label{fig:artic_box}
  \end{subfigure}
  \caption{Distribution of unique word count, mean utterance length and articulation rate, grouped by RI and language.}
  \label{fig:box_speech_prof}
\end{figure}

While PFI shows the importance of each feature in the context of all others, 
single-feature box plots help visualise how individual features relate to the target variable (RI) in isolation~\cite{williamson_parker_etal1989}.
Figure~\ref{fig:box_speech_prof} shows the median, quartiles, outliers and means (dotted lines) for selected features, grouped by intervention status: red (left) for RI and blue (right) for non-RI.
The trends in the box plots reinforce the findings of the PFI plots for (a) unique words and (b) mean utterance length, with higher values linked with typical development.
These 
plots also give insight into the surprising articulation rate results (c).
Among isiXhosa-speaking children, both RI and non-RI children have similar articulation rates.
In contrast, for Afrikaans, non-RI children (blue) have a higher median and maximum articulation rate, but the variance among this group is very large, spanning that of the RI group.

\textbf{Qualitative analysis:}
For both languages, unique word count is the strongest predictor of satisfactory development. 
This aligns with established findings~\cite{chan_chen_etal2023}, as unique word count measures lexical diversity and reflects a larger vocabulary, which is associated with more complex narratives. 
Similarly, mean utterance length is an important feature in both models. 
Utterance length captures syntactic complexity, a microstructural feature related to overall narrative macrostructure. 
Children requiring intervention typically produce descriptive and action sequences marked by short, simple sentences, often limited to single words or phrases. 
Consequently, shorter utterances are consistently associated with not meeting age-level expectations.
Surprisingly, a higher articulation rate is associated with a higher probability of requiring intervention for both groups. 
This may suggest that children who produced shorter narratives, often consisting of words, phrases, or simple sentences, spoke at a faster rate. 
In contrast, more complex narratives, involving more sophisticated syntax, may require a slower speech rate to allow for greater planning and linguistic processing.

\subsection{Which grammatical features are important?}
\label{sec:grammatical}
\hspace{\parindent}\textbf{Setup:}
Next we consider the relative importance of grammatical features~\cite{hengeveld_rijkhoff_etal2004}. E.g., are verbs or nouns more important in determining whether a child requires intervention?
Concretely, we derive part-of-speech (POS) counts for both the Afrikaans and isiXhosa data
using ctextcore,\footnote{\url{https://pypi.org/project/ctextcore/}}
an open-source Python package that provides corpus analysis tools for ten South African languages using universal POS tags~\cite{petrov_das_etal2012}.
We select a subset of seven
POS tags that apply to both languages. 
These include counts of verbs, nouns, pronouns, adverbs and adjectives. 
Additionally, we include auxiliary (helping) verbs and particles. 
Afrikaans particles include \textit{te} used for introducing an infinitive, the negation particle \textit{nie} and the genitival particle \textit{se}.

\begin{figure}[t!]
  \centering
  \begin{subfigure}[b]{0.49\linewidth}
    \includegraphics[width=\linewidth]{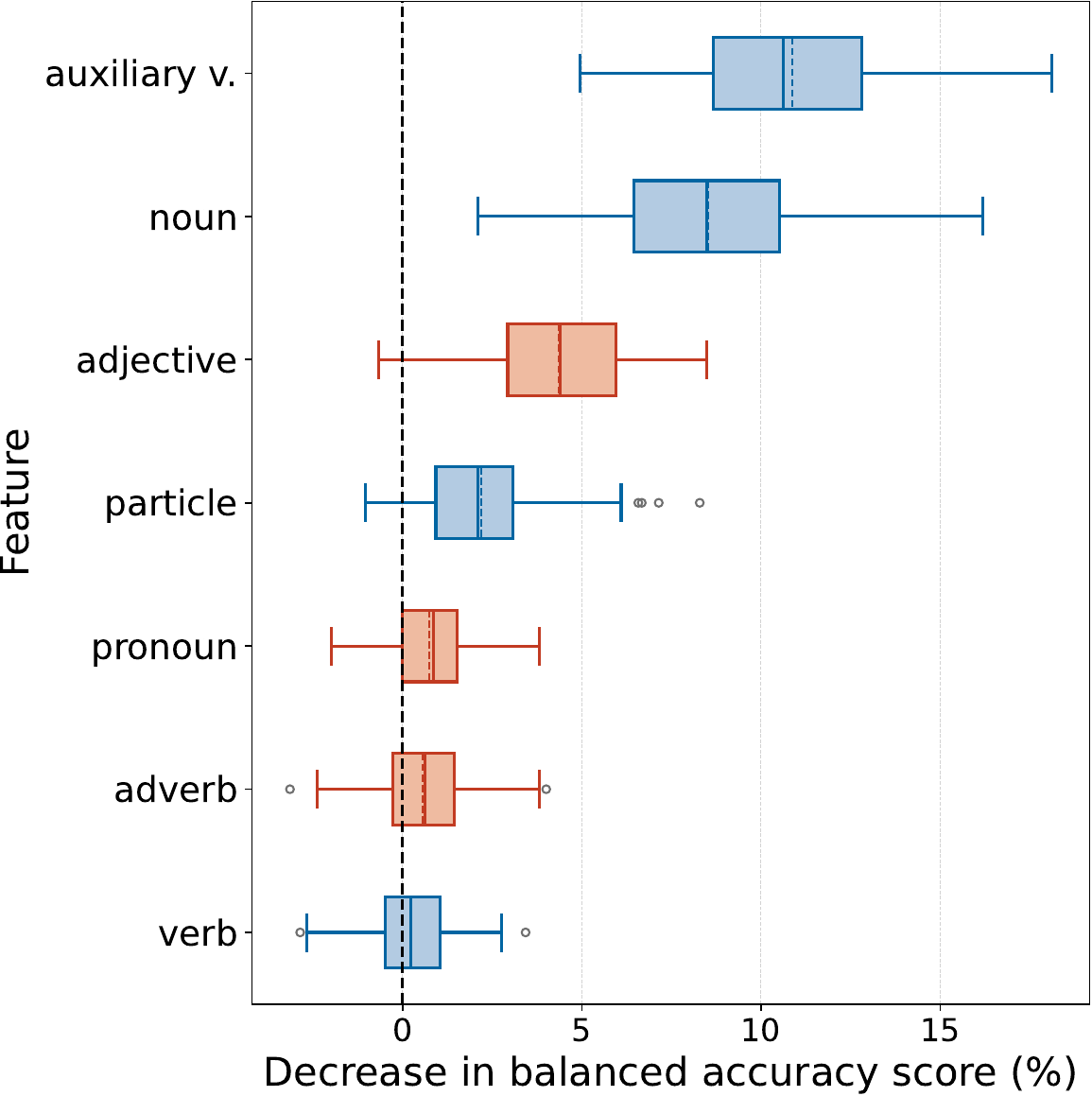}
    \caption{Afrikaans}
    \label{fig:PFI_af_UPOS}
  \end{subfigure}
  \hfill
  \begin{subfigure}[b]{0.49\linewidth}
    \includegraphics[width=\linewidth]{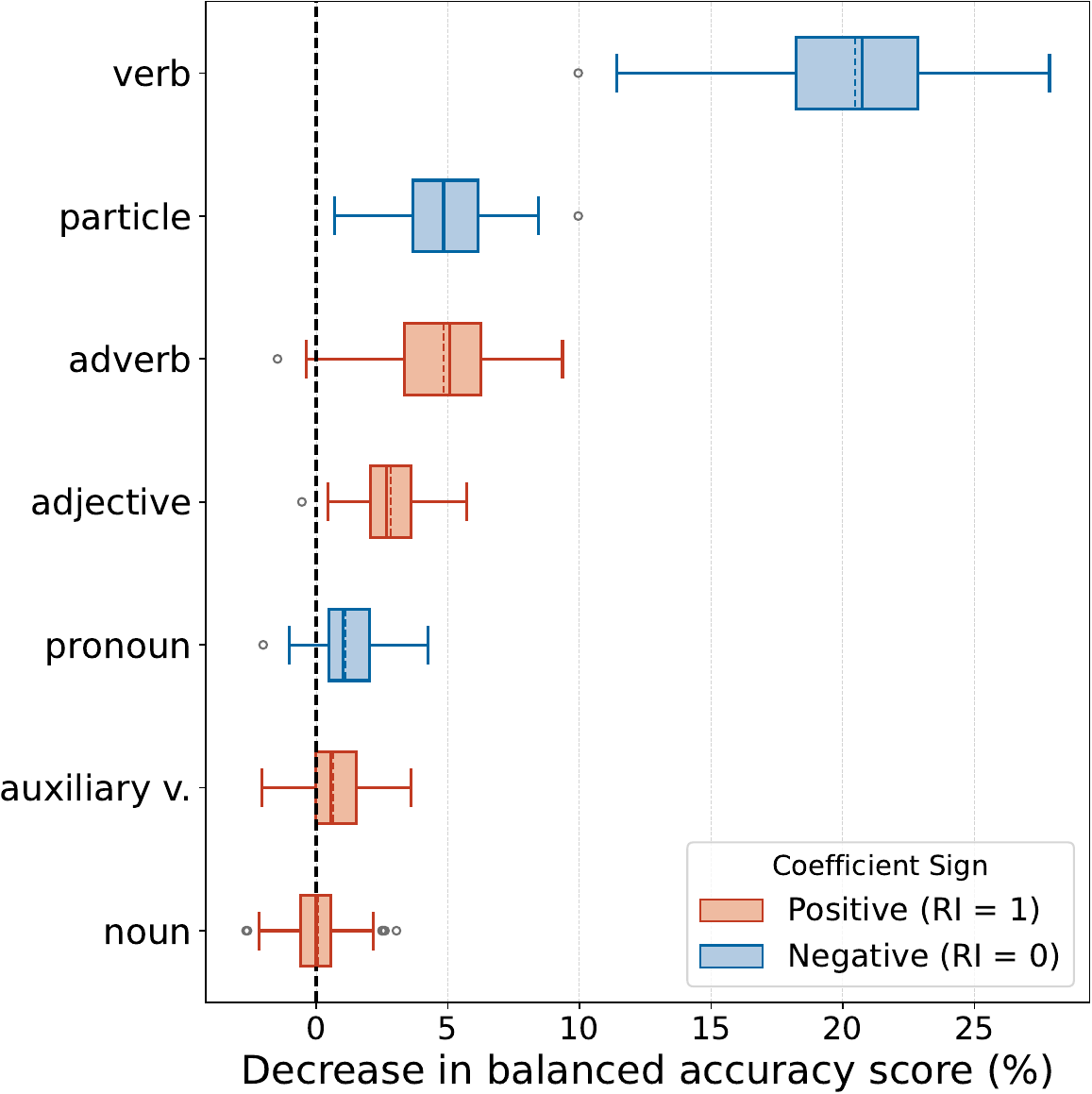}
    \caption{isiXhosa}
    \label{fig:PFI_xh_UPOS}
  \end{subfigure}
  \caption{PFI for grammatical features, showing the drop in balanced accuracy when a given feature is corrupted.}
  \label{fig:PFI_UPOS}
\end{figure}

\begin{figure}[t!]
  \centering
  \begin{subfigure}[b]{0.29\linewidth}
    \includegraphics[width=\linewidth]{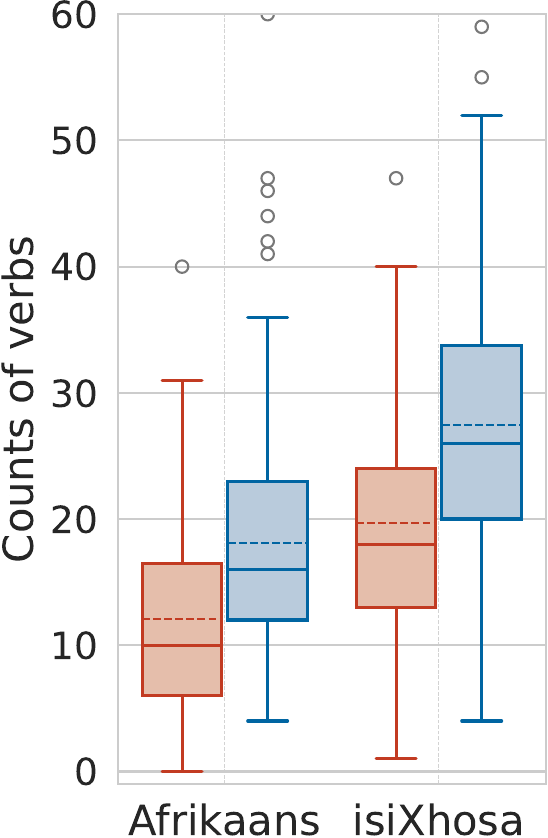}
    \caption{Verbs}
    \label{fig:box_verb}
  \end{subfigure}
  \begin{subfigure}[b]{0.29\linewidth}
    \includegraphics[width=\linewidth]{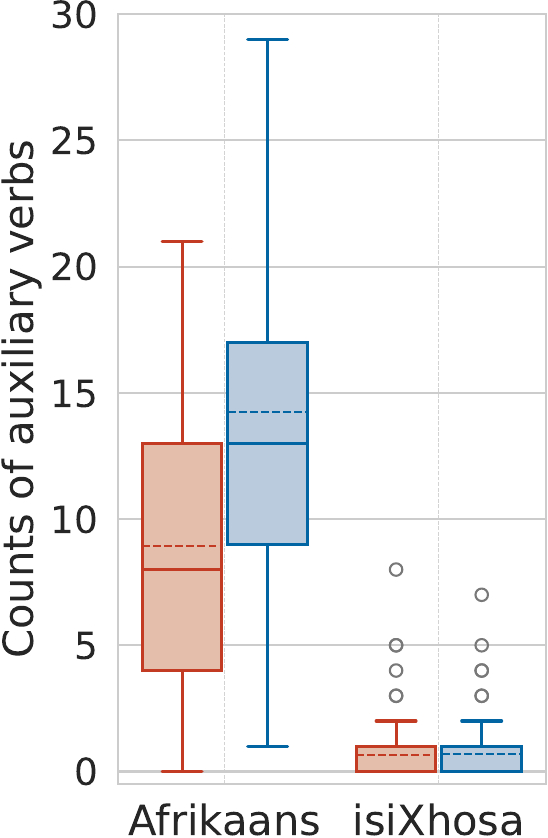}
    \caption{Auxiliary verbs}
    \label{fig:box_aux}
  \end{subfigure}
  \begin{subfigure}[b]{0.285\linewidth}
    \includegraphics[width=\linewidth]{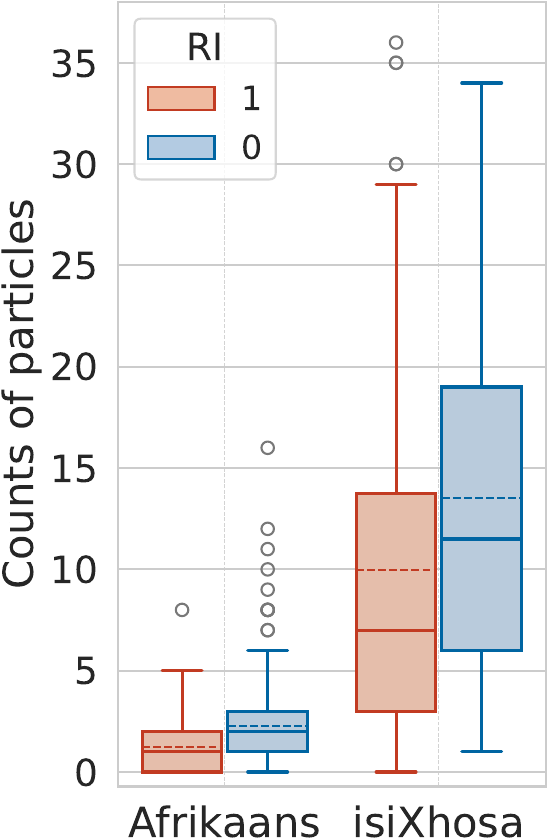}
    \caption{Particles}
    \label{fig:box_particle}
  \end{subfigure}
  \caption{Distribution of verbs, auxiliaries and particles in the text, grouped by whether intervention is required and language.}
  \label{fig:box_upos}
\end{figure}

\textbf{Quantitative analysis:}
Figure~\ref{fig:PFI_UPOS} presents the PFI analysis while Figure~\ref{fig:box_upos} shows box plots for individual features.
For isiXhosa, verbs are the most important, as seen in Figure~\ref{fig:PFI_xh_UPOS}.
This is mirrored in Figure~\ref{fig:box_verb}, with children who need intervention using fewer verbs.
Surprisingly, verbs do not demonstrate much importance in the Afrikaans model; instead, auxiliary verbs stand out in Figure~\ref{fig:PFI_af_UPOS}, as is also confirmed in Figure~\ref{fig:box_aux}.
Although particles are important in both languages in Figure~\ref{fig:PFI_UPOS}, Figure~\ref{fig:box_particle} shows that they appear infrequently in the Afrikaans group, with most transcripts containing fewer than three particles.
Finally, it is interesting that using more adjectives and adverbs is associated with intervention predictions in both languages.

\textbf{Qualitative analysis:}
In general, verbs,
adjectives, adverbs, auxiliary verbs and particles contribute to the elaboration of noun and verb phrases, indicating higher syntactic complexity and therefore more advanced language skills (or less need for intervention). 
The predictive role of verbs and particles in isiXhosa, and of auxiliaries and particles in Afrikaans, is linguistically plausible for identifying non-RI children.
Other patterns are more difficult to interpret. 

\subsection{Are specific keywords important?}
\label{sec:keywords}

\hspace{\parindent}\textbf{Setup:}
The last model consists of counts of individual keywords, 
with a control feature to account for differences between the cat and dog stories (Sec.~\ref{sec:data}).
We start with 
the top twenty most common words from each language. 
We then use L1 regularisation to select ten keywords from each language.
L1 regularisation has the effect of shrinking coefficients to zero, thereby selecting relevant features and excluding non-informative ones~\cite{james_witten_etal2023, jurafsky_martin2025}.
We explored morphological parsing for isiXhosa, replacing words with morphemes~\cite{moeng_reay_etal2021}, and using n-gram features in both languages, but this 
did not improve development performance.

\begin{figure}[t]
  \centering
  \begin{subfigure}[b]{0.49\linewidth}
    \includegraphics[width=\linewidth]{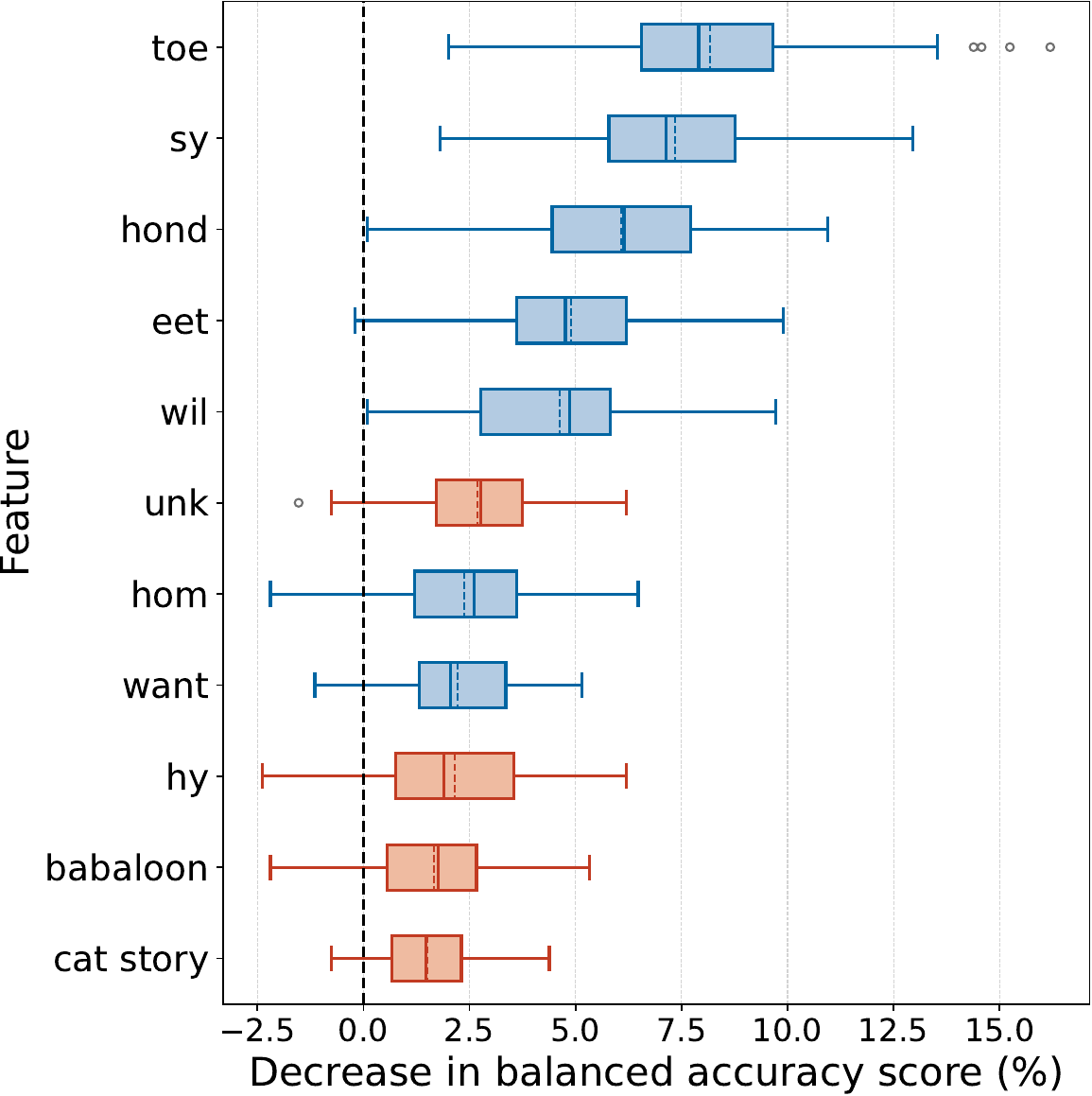}
    \caption{Afrikaans}
    \label{fig:PFI_af_words}
  \end{subfigure}
  \hfill
  \begin{subfigure}[b]{0.49\linewidth}
    \includegraphics[width=\linewidth]{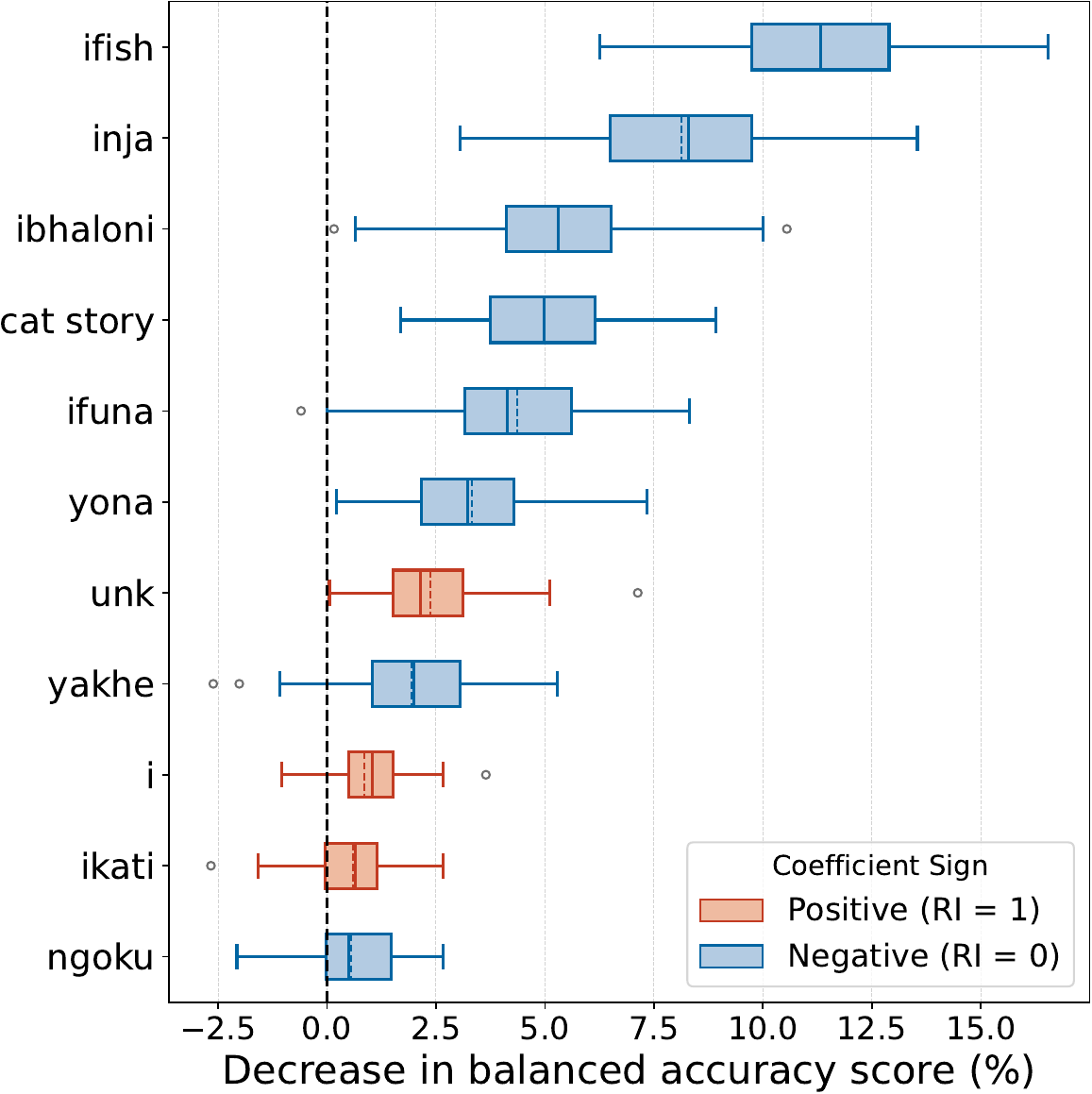}
    \caption{isiXhosa}
    \label{fig:PFI_xh_words}
  \end{subfigure}
  \caption{PFI for specific keywords, showing the drop in balanced accuracy when a given feature is corrupted.}
  \label{fig:PFI_words}
\end{figure}

\textbf{Quantitative analysis:}
As shown in Figure~\ref{fig:PFI_af_words}, the word \textit{toe} (then) has the highest feature importance in Afrikaans and is associated with the non-RI group (blue).
The pronoun \textit{sy}~(his), verb \textit{eet}~(eat) and auxiliary verb \textit{wil}~(want) also contribute meaningfully, with higher counts associated with a non-RI prediction.
Among the isiXhosa features, the auxiliary verb \textit{ifuna}~(want) and pronouns \textit{yona}~(it) and \textit{yakhe}~(his) are important features. 
It is interesting that the Afrikaans word \textit{wil} and the isiXhosa word \textit{ifuna} are important in both cases 
as they are both translations of the English word \textit{want}.
Nouns such as \textit{hond} (dog) in Afrikaans and \textit{ifish} (fish), \textit{inja} (dog) and \textit{ibhaloni} (balloon) in isiXhosa all demonstrate importance.
These nouns reflect key story elements, and their greater presence may signal longer, more developed narratives.
In Afrikaans (Figure~\ref{fig:PFI_af_words}), \textit{babaloon} has a positive coefficient (red), so the more it is used the higher the intervention prediction. 
The word \textit{babaloon} may occur more frequently in the intervention group because it is a mispronunciation of the Afrikaans word \textit{ballon} (balloon).
Similarly, we notice that the feature \textit{unk} (unknown), used to label unintelligible speech, is positively associated with intervention for both languages. 
This aligns with expectations, as unclear speech can be a sign of delays in oral narrative development.

\textbf{Qualitative analysis:}
The identification of \textit{toe} as a strong predictor in the Afrikaans model is well-founded; this adverb conveys both temporal progression and causality, which are typical of more advanced narrative structure.
Similarly, recognising \textit{wil} and \textit{ifuna} as predictors is reasonable, as these auxiliary verbs are often used in goal-oriented statements that reflect more structurally complex narratives.
The pronouns \textit{sy} in Afrikaans and \textit{yakhe} in isiXhosa may contribute as predictors because they elaborate noun phrases, which tend to occur in more syntactically complex constructions. 
The remaining patterns are less straightforward to interpret qualitatively.

\section{Summary and conclusion}
\label{sec:conclusion}

We have looked at the relative importance of verbal language proficiency, grammatical and keyword features in oral narratives from four- and five-year-olds.
In both our Afrikaans and isiXhosa data, lexical diversity and utterance length distinguished children needing intervention 
from those who do not.
Certain language-specific verbs, auxiliaries and nouns linked to core story elements were indicative of typical development, while unintelligible speech and mispronunciations were associated with requiring 
intervention. 
These findings are consistent with prior qualitative linguistic analyses reported in the literature.
However, some results challenged expectations:
speech production features such as articulation rate and grammatical elements like adjectives and adverbs showed low feature importance.
Rather than providing clear answers, these results raise important questions about the
role of these features in assessing early narrative skills.
Further work by educators and researchers is therefore required to clarify these patterns.
In summary, our
findings highlight the complexities of assessing early language development, and in particular identifying the salient features 
linked to the need for language intervention.
With continued investigation, we hope these insights can be translated into practical tools to support assessment practices 
in preschool classrooms. 

\newpage
\section{Acknowledgements} 
This work was supported by a grant from the Het Jan Marais Fonds (HJMF).\\

\bibliographystyle{IEEEtran}
\bibliography{mybib}

\end{document}